\documentclass[a4paper,twoside]{article}
\usepackage{epsfig}
\usepackage{subcaption}
\usepackage{calc}
\usepackage{amssymb}
\usepackage{amstext}
\usepackage{amsmath}
\usepackage{amsthm}
\usepackage{mathtools}
\usepackage{multicol}
\usepackage{pslatex}
\usepackage{apalike}
\usepackage{algorithm2e}
\usepackage{tikz}
\usepackage{graphicx}
\let\bibhang\undefined
\usepackage{natbib}

\usepackage[bottom]{footmisc}
\usepackage{SCITEPRESS}     

\usetikzlibrary{automata}
\usetikzlibrary{positioning, arrows.meta, fit}

\newtheorem{definition}{Definition}

\tikzstyle{startstop} = [rectangle, rounded corners, minimum width=1.5cm, minimum height=0.5cm,text centered, draw=black, fill=red!30]
\tikzstyle{io} = [trapezium, trapezium left angle=70, trapezium right angle=110, minimum width=1cm, minimum height=0.5cm, text centered, draw=black, fill=blue!30]

\tikzstyle{process} = [rectangle, minimum width=3cm, minimum height=0.5cm, text centered, draw=black, fill=orange!30]
\tikzstyle{decision} = [diamond, minimum width=0.5cm, minimum height=0.1cm, text centered, draw=black, fill=green!30]
\tikzstyle{process2} = [rectangle, minimum width=1cm, minimum height=0.5cm, text centered, draw=black, fill=orange!30]
\tikzstyle{arrow} = [thick,->,>=stealth]
\usetikzlibrary{shapes,shadows,arrows,positioning,angles,quotes}
\tikzset{My Arrow Style/.style={single arrow, fill=black!15, anchor=base, align=center,text width=2.3cm}}
\tikzstyle{arrow} = [thick,->,>=stealth]
\usetikzlibrary{calc,trees,positioning,arrows,fit,shapes,calc}

\begin{document}

\title{Probabilistic Model Checking of Stochastic Reinforcement Learning Policies}

\author{\authorname{Dennis Gross, and Helge Spieker}
\affiliation{Simula Research Laboratory, Norway}
\email{dennis@simula.no}
}


\abstract{We introduce a method to verify stochastic reinforcement learning (RL) policies.
This approach is compatible with any RL algorithm as long as the algorithm and its corresponding environment collectively adhere to the Markov property.
In this setting, the future state of the environment should depend solely on its current state and the action executed, independent of any previous states or actions.
Our method integrates a verification technique, referred to as model checking, with RL, leveraging a Markov decision process, a trained RL policy, and a probabilistic computation tree logic (PCTL) formula to build a formal model that can be subsequently verified via the model checker Storm.
We demonstrate our method's applicability across multiple benchmarks, comparing it to baseline methods called deterministic safety estimates and naive monolithic model checking.
Our results show that our method is suited to verify stochastic RL policies.}

\onecolumn \maketitle \normalsize \setcounter{footnote}{0} \vfill

\section{\uppercase{Introduction}}
\label{sec:introduction}
\emph{Reinforcement Learning (RL)} has revolutionized the industry, enabling the creation of agents that can outperform humans in sequential decision-making tasks~\citep{DBLP:journals/corr/MnihKSGAWR13,DBLP:journals/nature/SilverHMGSDSAPL16,DBLP:journals/nature/VinyalsBCMDCCPE19}.

In general, an RL agent aims to learn a near-optimal policy to achieve a fixed objective by taking actions and receiving feedback through rewards and observations from the environment~\citep{sutton2018reinforcement}.
We call a policy \emph{memoryless policy} if it only decides based on the current observation.
A \emph{deterministic policy} selects the same action in response to a specific observation every time, whereas a \emph{stochastic policy} might select different actions when faced with the same observation.
A popular choice is to employ a function approximator like a \emph{neural network (NN)} to model such policies.

\paragraph{RL problem.}
Unfortunately, learned policies are not guaranteed to avoid \emph{unsafe behavior}~\citep{DBLP:journals/jmlr/GarciaF15,DBLP:conf/aaai/Carr0JT23,DBLP:conf/aaai/AlshiekhBEKNT18}.
Generally, rewards lack the expressiveness to encode complex safety requirements \citep{littman2017environment,DBLP:conf/tacas/HahnPSSTW19,DBLP:conf/formats/HasanbeigKA20,DBLP:journals/aamas/VamplewSKRRRHHM22}.

To resolve the issues mentioned above, verification methods like \emph{model checking}~\citep{DBLP:books/daglib/0020348} are used to reason about the safety of RL, see for instance \citep{yuwangPCTL,DBLP:conf/formats/HasanbeigKA20,DBLP:conf/atva/BrazdilCCFKKPU14,DBLP:conf/tacas/HahnPSSTW19}.
Model checking is not limited by properties that can be expressed by rewards but support a broader range of properties that can be expressed by \emph{probabilistic computation tree logic} \citep[PCTL;][]{DBLP:journals/fac/HanssonJ94}.
At its core, model checking is a formal verification technique that uses mathematical models to verify the correctness of a system concerning a given (safety) property.

\paragraph{Hard to verify.}
While existing research focuses on verifying stochastic RL policies~\citep{DBLP:conf/formats/Bacci020,DBLP:conf/ijcai/BacciG021,DBLP:conf/nfm/BacciP22}, these existing verification methods do not scale well with neural network policies that contain many layers and neurons.

\paragraph{Approach.}
This paper presents a method for verifying memoryless stochastic RL policies independent of the number of NN layers, neurons, or the specific memoryless RL algorithm.

Our approach hinges on three inputs: a Markov Decision Process (MDP) modeling the RL environment, a trained RL policy, and a probabilistic computation tree logic (PCTL) formula~\citep{DBLP:journals/fac/HanssonJ94} specifying the safety measurement.
Using an incremental building process~\citep{DBLP:conf/setta/GrossJJP22,DBLP:conf/concur/CassezDFLL05,DBLP:conf/tacas/DavidJLMT15} for the formal verification, we build only the reachable MDP portion by the trained policy.
Utilizing Storm as our model checker~\citep{DBLP:journals/sttt/HenselJKQV22}, we assess the policy's safety measurement, leveraging the constructed model and the PCTL formula.

Our method is evaluated across various RL benchmarks and compared to an alternative approach that only builds the part of the MDP that is reachable via the highest probability actions and an approach called naive monolithic model checking.
The results confirm that our approach is suitable for verifying stochastic RL policies.

\section{\uppercase{Related Work}}
There exist a variety of related work focusing on the trained RL policy verification~\citep{DBLP:conf/sigcomm/EliyahuKKS21,DBLP:conf/sigcomm/KazakBKS19,pmlr-v161-corsi21a,DBLP:journals/corr/DragerFK0U15,DBLP:conf/pldi/ZhuXMJ19,DBLP:conf/seke/JinWZ22}.
\cite{DBLP:conf/formats/Bacci020} propose MOSAIC (MOdel SAfe Intelligent Control), which combines abstract interpretation and probabilistic verification to establish probabilistic guarantees of safe behavior.
The authors model the system as a continuous-space discrete-time Markov process and focus on finite-horizon safety specifications.
They tackle the challenges of infinite initial configurations and policy extraction from NN representation by constructing a finite-state abstraction as an MDP and using symbolic analysis of the NN.
The \emph{main difference} to our approach lies in constructing the induced discrete-time Markov chain (DTMC).
Our approach verifies a specific probabilistic policy derived from an MDP.
In contrast, the MOSAIC approach verifies safety guarantees by creating a finite-state abstraction of the MDP that accommodates the NN-based policy and focuses on finding safe regions of initial configurations.
The abstraction must also extract the policy from its NN representation.
Our approach only queries the trained policy and is not limited by the time horizon.

\cite{DBLP:conf/ijcai/BacciG021} presented the first technique for verifying if a NN policy controlling a dynamical system maintains the system within a safe region for an unbounded time.
The authors use template polyhedra to overapproximate the reach set and formulate the problem of computing template polyhedra as an optimization problem.
They introduce a MILP (Mixed-Integer Linear Programming) encoding for a sound abstraction of NNs with ReLU activation functions acting over discrete-time systems.
Their method supports linear, piecewise linear, and non-linear systems with polynomial and transcendental functions.
The safety verification verifies agents against a model of the environment.
The main difference to our approach is that we are independent of the memoryless policy architecture and do not need to encode the verification problem as a MILP problem.

\cite{DBLP:conf/nfm/BacciP22} define a formal model of policy execution using continuous-state, finite-branching discrete-time Markov processes and build and solve sound abstractions of these models.
The paper proposes a new abstraction based on interval Markov decision processes (IMDPs) to address the challenge of probabilistic policies specifying different action distributions across states.
The authors present methods to construct IMDP abstractions using template polyhedra and MILP for reasoning about the NN policy encoding and the RL agent's environment.
Furthermore, they introduce an iterative refinement approach based on sampling to improve the precision of the abstractions.

\cite{DBLP:conf/setta/GrossJJP22} incrementally build a formal model of the trained RL policy and the environment.
They then use the model checker Storm to verify the policy's behavior.
They support memoryless stochastic policies by always choosing the action with the highest probability (deterministic safety estimation).
In comparison, we build the model based on all possible actions with a probability greater than zero.

\section{Background}
This section describes probabilistic model checking and investigates RL's details.

\subsection{Probabilistic Model Checking}
A \textit{probability distribution} over a set $X$ is a function $\mu \colon X \rightarrow [0,1]$ with $\sum_{x \in X} \mu(x) = 1$. The set of all distributions on $X$ is denoted $Distr(X)$.

\begin{definition}[Markov Decision Process]\label{def:mdp}
A \emph{Markov decision process (MDP)} is a tuple $M = (S,s_0,Act,Tr, rew,AP,L)$ where
$S$ is a finite, nonempty set of states; $s_0 \in S$ is an initial state; $Act$ is a finite set of actions; $Tr\colon S \times Act \rightarrow Distr(S)$ is a partial probability transition function;
$rew \colon S \times Act \rightarrow \mathbb{R}$ is a reward~function;
$AP$ is a set of atomic propositions;
$L \colon S \rightarrow 2^{AP}$ is a labeling function that assigns atomic propositions to states.
\end{definition}
We employ a factored state representation where each state $s$ is a vector of features $(f_1, f_2, ...,f_d)$ where each feature $f_i\in \mathbb{Z}$ for $1 \leq i \leq d$ ($d$ is the dimension of the state).
Furthermore, we denote $Tr(s, a)(s')$ with $Tr(s, a, s')$ and $Tr(s,a,s')$ can be written as $s \xrightarrow{a} s'$.
If $\mid Act(s) \mid = 1$, we can omit the action $a$ in $Tr(s,a,s')$ and write $Tr(s,s')$.
Also, we define $NEIGH(s)$ for the set of all states $s' \in S$ that $Tr(s,s') \neq 0$.

The available actions in $s \in S$ are $Act(s) = \{a \in Act \mid Tr(s,a) \neq \bot\}$ where $Tr(s, a) \neq \bot$ is defined as action $a$ at state $s$ does not have a transition (action $a$ is not available in state $s$).
An MDP with only one action per state ($\forall s \in S : \mid Act(s)\mid  = 1$) is a DTMC $D$.

\begin{definition}[Discrete-time Markov Chain]
A discrete-time Markov chain (DTMC) is an MDP such that $\mid Act(s)\mid  = 1$ for all states $s \in S$. We denote a DTMC as a tuple $M_D = (S, s_0, Tr, rew)$ with $S, s_0, rew$ as in Definition~\ref{def:mdp} and transition probability function $Tr: S \rightarrow Distr(S)$.
\end{definition}

When an agent is executed in an environment modeled as an MDP, a \emph{policy} is a rule that an agent follows in deciding which action to take based on its current observation of the state of the environment to optimize its objective.

\begin{definition}[Deterministic Policy]
A memoryless deterministic policy for an MDP $M {=} (S,s_0,Act,Tr,rew)$ is a function $\pi \colon S \rightarrow Act$ that maps a state $s \in S$ to action $a \in Act$.
\end{definition}
Applying a policy $\pi$ to an MDP $M$ yields an \emph{induced DTMC} $D$ where all non-determinisms within the system are resolved.

\begin{definition}[Stochastic Policy]
A memoryless stochastic policy for an MDP $M {=} (S,s_0,Act,T,rew)$ is a function $\hat{\pi} \colon S \rightarrow Distr(Act)$ that maps a state $s \in S$ to distribution over actions $Distr(Act)$.
\end{definition}

Permissive policies facilitate a richer exploration of the state space by allowing the selection of multiple actions at each observation, potentially uncovering nuanced policies.

\begin{definition}[Permissive Policy] A permissive policy $\tau \colon S \rightarrow 2^{Act}$ selects multiple actions in every state.
\end{definition}

We specify the properties of a DTMC via the specification language PCTL~\citep{yuwangPCTL}.
\begin{definition}[PCTL Syntax]\label{def:pctl}
    Let $AP$ be a set of atomic propositions. The following grammar defines a state formula:
    $\Phi  ::= \text{ true }\mid \text{ a }\mid  \text{ } \Phi_1 \land \Phi_2 \text{ }\mid \text{ }\lnot \Phi \mid P_{\bowtie p}\mid  P^{max}_{\bowtie p}(\phi)\text{ }\mid \text{ }P^{min}_{\bowtie p}(\phi)$ where $a \in AP, \bowtie \in \{<,>,\leq,\geq\}$, $p \in [0,1]$ is a threshold, and $\phi$ is a path formula which is formed according to the following grammar $$\phi ::= X\Phi\text{ }\mid \text{ }\phi_1\text{ }U\text{ }\phi_2\text{ }\mid \text{ }\phi_1\text{ }F_{\theta_i t}\text{ }\phi_2\text{ } \text{ with }\theta_i = \{<,\leq\}.$$
    We further define ``eventually'' as $\lozenge \phi \coloneqq true \text{ U } \phi$.
\end{definition}
For MDPs, PCTL formulae are interpreted over the states of the induced DTMC of an MDP and a policy.
In a slight abuse of notation, we use PCTL state formulas to denote probability values. That is, we sometimes write $P_{\bowtie p}(\phi)$ where we omit the threshold $p$. 
For instance, in this paper, $P(\lozenge coll)$  denotes the reachability probability of eventually running into a collision.
There exist a variety of model checking algorithms for verifying PCTL properties~\citep{DBLP:conf/focs/CourcoubetisY88,DBLP:journals/jacm/CourcoubetisY95}.
PRISM~\citep{DBLP:conf/cav/KwiatkowskaNP11} and Storm~\citep{DBLP:journals/sttt/HenselJKQV22} offer efficient and mature tool support for verifying probabilistic systems.

\subsection{Reinforcement Learning}
\begin{figure}[]
\centering
\scalebox{1}{
    \begin{tikzpicture}[]
     {};
    \node (agent1) [process] {RL Agent};
    \node (env) [process, below of=agent1,yshift=-0.25cm,xshift=2cm] {Environment};
    
    \draw [arrow] (agent1) -| node[anchor=west] {Action} (env);
    \draw [arrow] (env) -| node[anchor=east] {New State, Reward} (agent1);
    \end{tikzpicture}
}
\caption{This diagram represents an RL system in which an agent interacts with an environment. The agent receives a state and a reward from the environment based on its previous action. The agent then uses this information to select the next action, which it sends to the environment.}
\label{fig:rl}
\end{figure}
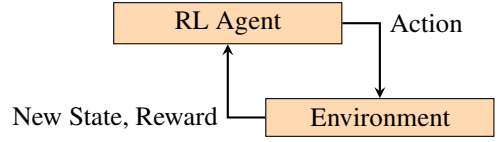
For MDPs with many states and transitions, obtaining optimal policies $\pi^*$ is difficult.
The standard learning goal for RL is to learn a policy $\pi$ in an MDP such that $\pi$ maximizes the accumulated discounted reward, that is, $\mathbb{E}[\sum^{N}_{t=0}\gamma^t R_t]$, where $\gamma$ with $0 \leq \gamma \leq 1$ is the discount factor, $R_t$ is the reward at time $t$, and $N$ is the total number of steps (see Figure~\ref{fig:rl}).
In RL, an agent learns through interaction with its environment to maximize a reward signal, via RL algorithm such as deep Q-learning~\citep{mnih2013playing}.
Note that rewards lack the expressiveness to encode complex safety requirements which can be specified via PCTL~\citep{littman2017environment,DBLP:conf/tacas/HahnPSSTW19,DBLP:conf/formats/HasanbeigKA20,DBLP:journals/aamas/VamplewSKRRRHHM22}.
Therefore, model checking is needed to verify that the trained RL policies satisfy the complex requirements.

\section{Methodology}
In this section, we introduce a method for verifying the safety of stochastic RL policies via probabilistic model checking concerning a safety measurement.
Given the MDP $M$ of the environment, a trained RL policy $\hat{\pi}$, and a safety measurement $m$, the general workflow is as follows and will be detailed below:

\begin{enumerate}
    \item \textbf{Induced MDP Construction}: We construct a new MDP using the original MDP and the trained RL policy \( \hat{\pi} \). This new MDP includes only the states that are reachable by the policy and actions that the policy considers, i.e., those for which the policy selection probability is greater than zero.

    \item \textbf{Induced DTMC Transformation}: The newly constructed MDP $\hat{M}$ is transformed into an induced DTMC, denoted as \( \hat{D} \). The transition probabilities between states in $\hat{D}$ are updated based on both the modified MDP \( \hat{M} \) and the action distribution of the original RL policy \( \hat{\pi} \).

    \item \textbf{Safety Verification}: Finally, the safety measurement $m$ of $\hat{D}$ is rigorously verified using the Storm model checker.
\end{enumerate}

\paragraph{Running example.} To elucidate the methodology, let's consider a running example featuring an MDP that models a grid-like environment (see Figure~\ref{subfigure:mdp_example1}). In this environment, an RL agent can navigate among four discrete states, denoted as \(x=1\), \(x=2\), \(x=3\), and \(x=4\), corresponding to positions \(A\), \(B\), \(C\), and \(D\), respectively. From each state, the agent can execute one of three actions: \( \text{UP} \), \( \text{NOP} \) (No Operation), or \( \text{DOWN} \).
The transition probabilities between states are defined based on the chosen actions.
For instance, executing the \( \text{UP} \) action from state \(A\) (\(x=1\)) gives the agent a \(0.2\) probability of transitioning to state \(B\) (\(x=2\)) and a \(0.8\) probability of transitioning to state \(C\) (\(x=3\)).
For simplicity, let's assume one reward for each state-action pair.
This example functions as a foundational MDP against which we will test our verification method.

\subsection{Induced MDP Construction}
To perform the verification of stochastic RL policies, we initially construct an induced MDP \( \hat{M} \) from a given MDP \( M \) (as shown in Figure~\ref{subfigure:mdp_example1}), a trained stochastic RL policy \( \hat{\pi} \), and a safety measurement \( m \).

We commence the construction process at the initial state \( s_0 \) of the original MDP \( M \).
Starting from the initial state \( s_0 \), we iteratively visit each state \( s \) that is reachable under the policy \( \hat{\pi} \) by only including actions \( a \) for which \( \hat{\pi}(a|s) > 0 \).
This creates an MDP \( \hat{M} \) induced by a permissive policy $\tau$ and ensures that \( \hat{M} \) is reduced to those states and actions that are reachable by \( \hat{\pi} \) (see Figure~\ref{fig:combined_actions}).
This induced MDP serves as the basis for the subsequent transformation into an induced DTMC \( \hat{D} \) for formal verification.

\begin{figure}[th!]
    \begin{minipage}{0.48\textwidth}
        \centering
        \begin{tikzpicture}[->, >=stealth', auto, semithick, node distance=1.5cm]
    \tikzstyle{every state}=[fill=white,draw=black,thick,text=black,scale=1, shape=ellipse, minimum width=0.5cm, minimum height=0.5cm]
    \node[state]    (A)                    {\tiny A : $x=1$};
    \node[right of=A](A_UP){\tiny UP};
    \node[below of=A_UP,yshift=1cm](A_DOWN){\tiny DOWN};
    \node[state]    (B)[right of=A_UP]   {\tiny B : $x=2$};

    \node[state]    (C)[below of=B]   {\tiny C : $x=3$};
    \node[state]    (D)[below of=A]   {\tiny D : $x=4$};

    \path
    (A) edge[right]     node[below right]{}     (A_UP)
    (A) edge[right]     node[below right]{}     (A_DOWN);

    \path
    (A) edge[loop above]     node{\tiny UP,NOP,DOWN}         (A);
    
    \path
    (A_UP) edge[left] node[below]{\tiny $0.2$} (B)
    (A_UP) edge[right] node[right]{\tiny $0.8$} (C);

    \path
    (A_DOWN) edge[left] node[below]{\tiny $0.4$} (C)
    (A_DOWN) edge[right] node[below]{\tiny $0.6$} (D);

    \path
    (B) edge[loop below]     node[right,xshift=0.5em]{\tiny UP,NOP,DOWN}         (B);

    \path
    (C) edge[loop below]     node[below,yshift=-1em,xshift=0.5em]{\tiny UP,NOP,DOWN}         (C);

    \path
    (D) edge[loop below]     node{\tiny UP,NOP,DOWN}         (D);
    
    \end{tikzpicture}
        \caption{The MDP $M$ with $S = \{x=1,x=2,x=3,x=4\}$, A=\{UP,NOP,DOWN\}, and $rew \colon S \times A \rightarrow \{1\}$.}
        \label{subfigure:mdp_example1}
    \end{minipage}

    \vspace{1em}

    \begin{minipage}{0.48\textwidth}
        \centering
        \begin{tikzpicture}[->, >=stealth', auto, semithick, node distance=1.5cm]
    \tikzstyle{every state}=[fill=white,draw=black,thick,text=black,scale=1, shape=ellipse, minimum width=0.5cm, minimum height=0.5cm]
    \node[state, draw=blue, text=blue]    (A)                    {\tiny A : $x=1$};
    \node[right of=A](A_UP){\tiny UP};
    \node[below of=A_UP,yshift=1cm](A_DOWN){\tiny DOWN};
    \node[state]    (B)[right of=A_UP]   {\tiny B : $x=2$};

    \node[state]    (C)[below of=B]   []{\tiny C : $x=3$};
    \node[state]    (D)[below of=A]   {\tiny D : $x=4$};

    \path
    (A) edge[right]     node[below right]{}     (A_UP)
    (A) edge[right]     node[below right]{}     (A_DOWN);

    \path
    (A_UP) edge[left] node[below]{\tiny $0.2$} (B)
    (A_UP) edge[right] node[right]{\tiny $0.8$} (C);

    \path
    (A_DOWN) edge[left] node[below]{\tiny $0.4$} (C)
    (A_DOWN) edge[right] node[below]{\tiny $0.6$} (D);

    (B) edge[loop below]     node[right,xshift=0.5em]{\tiny UP,NOP}         (B);


    
    \end{tikzpicture}
        \caption{Induced MDP $\hat{M}$. We add the state-action transition to the MDP for each action with a probability greater than zero $\hat{\pi}(a|s)$. In this example, the chosen actions are UP and DOWN with the corresponding probabilities $\hat{\pi}(UP|x=1) = 0.3$ and $\hat{\pi}(DOWN|x=1) = 0.7$ at state $x=1$.}
        \label{fig:combined_actions}
    \end{minipage}

    \vspace{1em}
    
    \begin{minipage}{0.48\textwidth}
        \centering
        \begin{tikzpicture}[->, >=stealth', auto, semithick, node distance=1.5cm]
    \tikzstyle{every state}=[fill=white,draw=black,thick,text=black,scale=1, shape=ellipse, minimum width=0.5cm, minimum height=0.5cm]
    \node[state, draw=blue, text=blue]    (A)                    {\tiny A : $x=1$};
    \node[right of=A](A_UP){};
    \node[below of=A_UP,yshift=1cm](A_DOWN){};
    \node[state]    (B)[right of=A_UP]   {\tiny B : $x=2$};

    \node[state]    (C)[below of=B]   {\tiny C : $x=3$};
    \node[state]    (D)[below of=A]   {\tiny D : $x=3$};

    \path
    (A) edge[right]     node[below right]{}     (A_DOWN);

    \path
    (A_DOWN) edge[left] node[above,xshift=-1.5em]{\tiny $0.3  \cdot 0.2$} (B)
    (A_DOWN) edge[left] node[above,yshift=-0.5em,xshift=2.4em]{\tiny $0.7 \cdot 0.4 + 0.3 \cdot 0.8$} (C)
    (A_DOWN) edge[right] node[below,yshift=-0.4em]{\tiny $0.7 \cdot 0.6$} (D);



    
    \end{tikzpicture}
        \caption{Transitions update for $\hat{D}$. For example, $\hat{Tr}_{\hat{D}}(x=1)(x=2) = Tr(x=1,UP,x=2) \cdot \hat{\pi}(UP\mid x=1) = 0.2 \cdot 0.3 = 0.06$ at state $x=1$. Repeat with step from Figure~\ref{fig:combined_actions} for all reachable states by the trained RL policy.}
        \label{fig:updated}
    \end{minipage}
    
\end{figure}

\subsection{Induced DTMC Transformation}
To convert the induced MDP \( \hat{M} \) into an induced DTMC \( \hat{D} \), we perform a probabilistic transformation on the state transitions (refer to Figure \ref{fig:updated}). Specifically, we update the transition function \( \hat{Tr}_{\hat{D}}(s, s') \), which defines the probability of transitioning from state \( s \) to state \( s' \), for each \( s' \) that is a neighboring state \( NEIGH(s) \) of \( s \) within the set of all states \( S_{\hat{M}} \) of the induced MDP \( \hat{M} \).

The transition function \( \hat{Tr}_{\hat{D}}(s, s') \) is given by the following equation:
\[
\hat{Tr}_{\hat{D}}(s,s') = \sum_{a \in Act_{\hat{M}}(s)} Tr_{\hat{M}}(s,a,s') \hat{\pi}(a|s)
\]
Where \( Act_{\hat{M}}(s) \) represents the set of actions available in state \( s \) of \( \hat{M} \), \( Tr_{\hat{M}}(s,a,s') \) denotes the original transition probability of moving from state \( s \) to state \( s' \) when action \( a \) is taken in \( \hat{M} \), and \( \hat{\pi}(a|s) \) is the action-selection probability of action \( a \) in state \( s \) under the RL policy \( \hat{\pi} \).
The equation essentially performs a weighted sum of all possible transitions from state \( s \) to state \( s' \), using the probabilities of selecting each action \( a \) in \( \hat{M} \) according to \( \hat{\pi} \) as the weights. This transformation inherently incorporates the decision-making behavior of the RL policy into the probabilistic structure of \( \hat{D} \), enabling us to perform verification tasks.

\subsection{Safety Verification}
Once the induced DTMC \( \hat{D} \) is constructed, the final step in our methodology is to do the safety measurement using the Storm model checker.
It takes as input the DTMC \( \hat{D} \) and a safety measurement $m$.
Upon feeding \( \hat{D} \) and the specified safety measurement $m$ into Storm, the model checker systematically explores all possible states and transitions in \( \hat{D} \) to return the safety measurement result.


\subsection{Limitations}
Our method is independent of the RL algorithm; therefore, we can verify any neural network size.
However, it requires the policy to obey the Markov property, i.e., there is no memory, and the action is selected only using the current state.
Furthermore, we are limited to discrete state and action spaces due to the discrete nature of the model checking component.

\section{Experiments}
We now evaluate our method in multiple RL environments.
We focus on the following research questions:
\begin{itemize}
    \item Can we model check RL policies in common benchmarks such as Freeway?
    \item How does our approach compare to COOL-MC's deterministic estimation and naive monolithic model checking?
    \item How does the method scale?
\end{itemize}
The experiments are performed by initially training the RL policies using the PPO algorithm~\citep{pytorch_minimal_ppo}, then using the trained policies to answer our research questions.

\begin{table}[t]
    \centering
    \scalebox{0.67}{
    \begin{tabular}{|lrrrr|}
        \hline
        \textbf{Environment} & \textbf{Layers} & \textbf{Neurons} & \textbf{Ep.} & \textbf{Reward} \\
        \hline
        Freeway & 2 & 64   & 898 & 	0.78\\
        Crazy Climber & 2 & 1024 &  3,229 &  76.22\\
        Avoidance & 2  & 64 & 7,749  & $6,194$  \\
        \hline
    \end{tabular}
    }
    \caption{Trained PPO Policies. The learning rates are $0.0001$, the batch sizes are $32$, and the seeds are 128. "Ep." is shorthand for "episode." The reward is averaged over a sliding window of 100 episodes.}
    \label{table:trained}
\end{table}

\begin{table*}[]
\centering
\scalebox{0.67}{
\begin{tabular}{|ll|rrrr|rrrr|rrrr|}
\hline
\multicolumn{2}{|l|}{\textbf{Setup}} & \multicolumn{4}{l|}{\textbf{Deterministic Policy}} & \multicolumn{4}{l|}{\textbf{Stochastic Policy}} & \multicolumn{4}{l|}{\textbf{Naive Monolithic (No RL Policy)}} \\
\textbf{Environment}  & \textbf{Measurement}    & \textbf{States}     & \textbf{Transitions} & \textbf{Result}    & \textbf{Time}    & \textbf{States}    & \textbf{Transitions}  & \textbf{Result}  & \textbf{Time}  & \textbf{States}    & \textbf{Transitions}  & \textbf{Result}  & \textbf{Time} \\
\hline
Freeway      & \(P(\lozenge \text{goal})\) & 123        & 340          & 0.7       & 2.5       & 496           & 2,000            & 0.7       & 11      & 496           & 2800            & 1.0       & 0.06 \\
Freeway      & \(P(\text{mid} \, \text{ U } \, \text{mid}_{-1})\) & 21        & 36          & 1.0       & 0.5       & 272           & 1,120            & 1.0       & 6        & 272           & 1552            & 1.0       & 0.07 \\
Crazy Climber      & \(P(\lozenge \text{coll})\) & \(8,192\)        & \(32,768\)          & 0.0       & 146       & 40,960           & 270,336              & 1.0       & 1,313       & 40,960           & 385,024            & 1.0       & 0.4 \\
Avoidance     & \(P(\lozenge_{\leq 100} \text{coll})\) & 625        & 8698          & 0.84       & 1.5       & 15,625           & 892,165              & 0.84       & 1,325         & 15,625           & 1,529,185            & 1.0       & 2.0 \\
\hline
\end{tabular}
}
\caption{RL benchmarks across various environments and metrics. Columns display the number of built states, transitions, safety measures, and computational time required. The time is calculated by adding the time taken to build the model to the time taken to check it and is expressed in seconds. Notably, the time for model checking is negligible and, therefore, not explicitly mentioned.}
\label{tab:benchmarks}
\end{table*}

\subsection{Setup}
We now describe our setup. First, we describe the used environments, then the trained policies, and finally, our technical setup.

\subsubsection{Environments.}
We focus on environments that have previously been used in the RL literature.

\paragraph{Freeway.} The RL agent controls a chicken (up, down, no operation) running across a highway filled with traffic to get to the other side.
Every time the chicken gets across the highway, it earns a reward of one.
An episode ends if the chicken gets hit by a car or reaches the other side.
Each state is an image of the game's state.
Note that we use an abstraction of the original game, which sets the chicken into the middle column of the screen and contains fewer pixels than the original game, but uses the same reward function and actions~\citep{mnih2015human}.

\paragraph{Avoidance.} This environment contains one agent and two moving obstacles in a two-dimensional grid world. The environment terminates when a collision between the agent and an obstacle happens. For each step that did not resolve in a collision, the agent gets rewarded with a reward of $100$. The environment contains a slickness parameter, which defines the probability that the agent stays in the same cell~\citep{DBLP:conf/setta/GrossJJP22}.
\paragraph{Crazy Climber.}
It is a game where the player has to climb a wall~\citep{mnih2015human}.
This environment is a PRISM abstraction based on this game.
Each state is an image.
A pixel with a One indicates the player's position.
A pixel with a Zero indicates an empty pixel.
A pixel with a Three indicates a falling object.
A pixel with a four indicates the player's collision with an object.
The right side of the wall consists of a window front.
The player must avoid climbing up there since the windows are unstable.
For every level the play climbs, the player gets a reward of 1.
To avoid falling obstacles, the player has the option to move left, move right, or stay idle.

\subsubsection{Trained RL policies}
We trained PPO agents in the previously introduced RL environments. The RL training results are summarized in Table~\ref{table:trained}.

\subsubsection{Technical setup}
We executed our benchmarks in a docker container with 16 GB RAM, and an AMD Ryzen 7 7735hs with Radeon graphics × 16 processor with the operating system Ubuntu 20.04.5 LTS.
For model checking, we use Storm 1.7.1 (dev)~\citep{DBLP:journals/sttt/HenselJKQV22}.

\subsection{Analysis}

We now go through our evaluation and answer our research questions.

\subsubsection{Can we verify stochastic RL policies in common benchmarks such as Freeway?}
In this experiment, we investigate the applicability of our approach to evaluating RL policies, specifically focusing on the Freeway environment. 

\paragraph{Execution.} Our analysis encompasses two dimensions: (1) the probability of reaching the desired goal, denoted by \(P(\lozenge \text{goal})\), and (2) the complex behavior involving the likelihood of the agent reverting to its starting position, denoted by \(P(\text{mid} \, U \, \text{mid}_{-1})\).

\paragraph{Results.} For the first dimension, we evaluated the probability of the trained Freeway RL agent successfully crossing the street without colliding with a car. The model checking result for this safety measurement yielded \(P(\lozenge \text{goal}) = 0.7\), indicating that the agent has a 70\% chance of safely reaching the other side of the road.

For the second dimension, we examined the agent's likelihood of reversing direction and returning to the starting position after having entered the street. The model checking process, in this case, revealed that \(P(\text{mid} \, \text{ U } \, \text{mid}_{-1}) = 1\), confirming that the agent will indeed return to the starting position with certainty under the analyzed conditions.

In summary, our findings demonstrate that it is not only feasible to apply model checking methods to stochastic RL policies in commonly used benchmarks like Freeway but also that these methods are versatile enough to evaluate complex safety requirements.

\subsubsection{How does our approach compare to COOL-MC's deterministic estimation and naive monolithic model checking?}
This experiment compares our approach with \emph{deterministic estimations of the safety measures of trained policies} and \emph{naive monolithic model checking}.
\emph{Deterministic estimation} incrementally builds the induced DTMC by querying the policy at every state for the highest probability action~\citep{DBLP:conf/setta/GrossJJP22}.
\emph{Naive monolithic model checking} is called "naive" because it does not take into account the complexity of the system or the number of possible states it can be in, and it is called "monolithic" because it treats the entire system as a single entity, without considering the individual components of the system or the interactions between them~\citep{DBLP:conf/aips/GrossS0023}.
In short, it does not verify the trained RL policy but rather the overall environment.
For instance, $P(\lozenge goal)=1$ indicates a path to reach the goal state eventually.

\paragraph{Execution.} We use the trained stochastic policy to build and verify the induced DTMCs correspondingly for the deterministic estimation and our approach.
For naive monolithic model checking, we build the whole MDP and verify it correspondingly.

\paragraph{Results.} The data presented in Table~\ref{tab:benchmarks} indicates that our method yields precise results (see Crazy Climber). In contrast, the deterministic estimation technique exhibits faster performance. This is due to its method of extending only the action transitions associated with the highest-probability action.
The naive monolithic model checking results are bounds and do not reflect the actual RL policy performance.
In the context of naive monolithic model checking, the number of states and transitions is larger than the other two approaches.
This indicates that naive monolithic model checking runs faster out of memory than the other two approaches, which is critical in environments with many states and transitions \citep{gross2023model}.

\subsubsection{How does the method scale?}
Based on our experiments, detailed in Table~\ref{tab:benchmarks}, we find that the primary limitations of our approach stem from the increasing number of states and transitions.
Given the probabilistic characteristics inherent in the RL policy, there is an increase in the number of states and transitions that can be reached compared to a deterministic RL policy.
This expansion results in more extended times for the overall model checking.
Optimizing our method's incremental building process may increase the model checking performance~\citep{DBLP:conf/aips/GrossS0023}.

\section{Conclusion}
We presented a methodology for verifying memoryless stochastic RL policies, thus addressing a gap in the current body of research regarding the safety verification of RL policies with complex, layered NNs.
Our method operates independently of the specific RL algorithm in use.
Furthermore, we demonstrated the effectiveness of our approach across various RL benchmarks, confirming its capability to verify stochastic RL policies comprehensively.

For future work, exploring the integration of safe RL~\citep{DBLP:conf/aaai/Carr0JT23} and stochastic RL verification offers a promising path to validate policies' reliability and enhance their operational safety across diverse environments.
Additionally, merging stochastic RL verification with interpretability~\citep{DBLP:conf/icmlc2/ZhaoDLZWW023} and explainability~\citep{DBLP:journals/csur/Vouros23} approaches could significantly bolster the understanding of RL policies.

\bibliographystyle{apalike}
{\small
\bibliography{example}}

\end{document}